\documentclass[10pt,journal,compsoc]{IEEEtran}

\ifCLASSOPTIONcompsoc
  \usepackage[nocompress]{cite}
\else
  \usepackage{cite}
\fi
\usepackage[pdftex]{graphicx}
\usepackage{multirow}
\usepackage[ruled,vlined]{algorithm2e}
\usepackage[numbers]{natbib}
\usepackage{xcolor}
\usepackage{amsfonts}
\usepackage{hyperref}

\definecolor{fred}{rgb}{0.188, 0.478, 0.858}
\definecolor{Gabe}{rgb}{0.988, 0.278, 0.258}
\definecolor{raquel}{RGB}{0,0,0}

\begin{document}

\title{Heterogeneous Multi-task Learning with Expert Diversity}

%
%

\author{Raquel~Aoki, 
        Frederick~Tung,
        ~and~Gabriel~L.~Oliveira,
\IEEEcompsocitemizethanks{\IEEEcompsocthanksitem R. Aoki is with the School of Computing Science, Simon Fraser University, Burnaby, Canada.\protect\\
E-mail: raoki@sfu.ca
\IEEEcompsocthanksitem F. Tung and G. Oliveira are with Borealis AI.}
\thanks{©2022 IEEE. Personal use of this material is permitted. Permission from IEEE must be obtained for all other uses, in any current or future media, including reprinting/republishing this material for advertising or promotional purposes, creating new collective works, for resale or redistribution to servers or lists, or reuse of any copyrighted component of this work in other works. Accepted article 2022.}}

%
%


\markboth{IEEE/ACM TRANSACTIONS ON COMPUTATIONAL BIOLOGY AND BIOINFORMATICS, Accepted 11 May 2022.}%
{Shell \MakeLowercase{\textit{et al.}}: IEEE/ACM TRANSACTIONS ON COMPUTATIONAL BIOLOGY AND BIOINFORMATICS}

%



\IEEEtitleabstractindextext{%
\begin{abstract}
Predicting multiple heterogeneous biological and medical targets is a challenge for traditional deep learning models. In contrast to single-task learning, in which a separate model is trained for each target, multi-task learning (MTL) optimizes a single model to predict multiple related targets simultaneously. To address this challenge, we propose the Multi-gate Mixture-of-Experts with Exclusivity (MMoEEx). Our work aims to tackle the heterogeneous MTL setting, in which the same model optimizes multiple tasks with different characteristics. Such a scenario can overwhelm current MTL approaches due to the challenges in balancing shared and task-specific representations and the need to optimize tasks with competing optimization paths. Our method makes two key contributions: first, we introduce an approach to induce more diversity among experts, thus creating representations more suitable for highly imbalanced and heterogenous MTL learning; second, we adopt a two-step optimization \cite{finn2017model, lee2020multitask} approach to balancing the tasks at the gradient level. We validate our method on three MTL benchmark datasets, including UCI-Census-income dataset, Medical Information Mart for Intensive Care (MIMIC-III) and PubChem BioAssay (PCBA).
\end{abstract}

}

\IEEEoverridecommandlockouts
\IEEEpubid{\makebox[\columnwidth]{978-1-5386-5541-2/18/\$31.00~\copyright2018 IEEE \hfill} \hspace{\columnsep}\makebox[\columnwidth]{ }}

\maketitle
\IEEEpubidadjcol

\IEEEdisplaynontitleabstractindextext

%
\IEEEpeerreviewmaketitle

\ifCLASSOPTIONcompsoc
\IEEEraisesectionheading{\section{Introduction}\label{sec:introduction}}
\else
\section{Introduction}
\label{sec:introduction}
\fi

\begin{figure*}[!ht]
\centering
  \includegraphics[width=\textwidth]{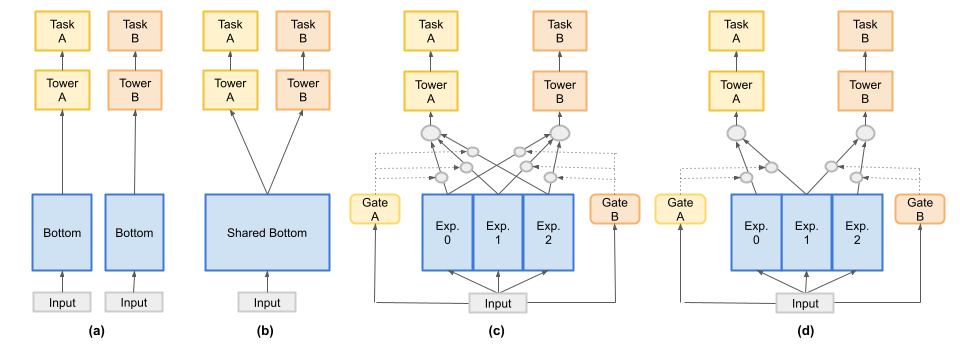}
  \caption{Neural Network Architectures.(a) Single-Task Learning, (b) Multi-task learning Hard-parameter sharing \cite{caruana1993multitask}, (c) Multi-gate Mixture-of-Experts (MMoE)\cite{ma2018modeling}, (d) MMoEEx -- proposed method.}
    \label{nnarchitectures}
\end{figure*}

\IEEEPARstart{S}{ingle} 
-task learning (STL) models are the most traditional approach in machine learning and have been extremely successful in many applications. This approach assumes that the model is required to output a single prediction target for a given input sample, such as a class label or a regression value. If two output targets are associated with the same input data, then two independent models are trained: one for each target or task (See Figure \ref{nnarchitectures}.a). STL may be suitable for situations in which the tasks are very different from each other, and in which computational efficiency may be ignored. However, when the tasks are related, STL models are parameter inefficient \cite{mallyaetal2018,zhaietal2020}.
In addition, in some applications, the synergy among tasks can help a jointly trained model better capture shared patterns that would otherwise be missed by independent training. For example, in computer vision, the synergy between the dense prediction tasks of semantic segmentation (the assignment of a semantic class label to each pixel in an image) and depth estimation (the prediction of real-world depth at each pixel in an image) can be leveraged to train a single neural network that achieves higher accuracy on both tasks than independently trained networks \cite{liuetal2019}.

In contrast to STL, multi-task learning (MTL) optimizes a single model to perform multiple related tasks simultaneously, aiming to improve generalization and parameter efficiency across tasks. In this case, two or more output targets are associated with the same input data (See Figures \ref{nnarchitectures}.b, \ref{nnarchitectures}.c and \ref{nnarchitectures}.d). Effective multi-tasking learning typically requires \textit{task balancing} to prevent one or more tasks from dominating the optimization, to decrease negative transfer, and to avoid overfitting. Standard MTL settings usually assume a homogeneous set of tasks, for example all tasks are classification or regression tasks, and usually they are non-sequential data. This scenario can greatly benefit MTL approaches with strong shared representations. 
In contrast, heterogeneous multi-task learning is defined by multiple classes of tasks, such as classification, regression with single or multi label characteristics and temporal data, being optimized simultaneously. The latter setting is more realistic but lacks further exploration. 
\citet{ma2018modeling} proposed a Multi-gate Mixture-of-Experts (MMoE), a model that combines \textit{experts} using gate functions. In this case, each expert is one or more neural network layers shared among the tasks. MMoE tends to generalize better than other models because it leverages several shared bottoms (experts) instead of using a single architecture. It allows dynamic parameter allocation to  shared and task-specific parts of the network thus improving further the representation power. In order to take advantage of these characteristics and extend it to heterogenous MTL problems, we introduced MMoEEx.       

The multi-gate mixture-of-experts with exclusivity (MMoEEx) model is a new mixture-of-experts (MMoE) approach to MTL that boosts the generalization performance of traditional MMoE via two key contributions:
\begin{itemize}
\item The experts in traditional MMoE are homogeneous, which limits the diversity of the learned representations.
Inspired by ensemble learning, we improve the generalization of traditional MMoE by inducing diversity among experts. We introduce novel exclusion and exclusivity conditions, under which some experts only contribute to some tasks, while other experts are shared among all tasks.
\item We introduce a two-step task balancing optimization at the gradient level inspired by MAML\cite{finn2017model}. This enables MMoEEx to support the learning of unbalanced heterogeneous tasks, in which some tasks may be more susceptible to overfitting, more challenging to learn, or operate at different loss scales.
\end{itemize}

We validate our proposed method with a broad set of experiments. First, we explored how MMoEEx behaves in the UCI Census-income dataset \cite{ma2018modeling,dua2019uci}, a standard MTL benchmark dataset for low cardinality tasks. We compare it with several state-of-the-art multi-task models and show that our technique outperforms the other approaches. Then, we evaluated the performance of MMoEEx on two medical and biological benchmark datasets. The Medical Information Mart for Intensive Care (MIMIC-III) \cite{johnson2016mimic, harutyunyan2019multitask} is a heterogeneous time series multi-task learning dataset. The mixture of multi-label and single-label temporal tasks with non-temporal binary classification makes this dataset ideal to benchmark our approach. The dataset's large-scale and high task imbalance characteristics also provide a scenario to exploit the robustness of our approach to competing tasks. We observed significant improvements in the AUC metrics against all compared approaches, especially the MMoE \cite{ma2018modeling} technique. Finally, we validated MMoEEx on the  PubChem BioAssay (PCBA) \cite{wang2017pubchem, liu2019loss} dataset. PCBA is a non-temporal homogeneous (only binary classification) high task cardinality dataset. While its tasks are less challenging than the MIMIC-III, it contains more than a hundred tasks. Thus, we adopt this dataset to benchmark the scalability and negative transfer aspects of MTL approaches. After exploring all these settings, our results confirmed the effectiveness of MMoEEx and showed that our approach has performance on par with the current state-of-the-art.

\section{Related Work}\label{related}

The recent works in deep learning for multi-task learning (MTL) can be divided into two groups: the ones focused on the neural network architecture, which study what, when and how to share information among the tasks; and the works focused on the optimization, which usually concentrate on how to balance competing tasks, which are jointly learned. Our work makes contributions to both fields.

MTL architectures can be divided into two main groups, hard parameter sharing and soft parameter sharing. One of the first works in MTL uses hard-parameter sharing. In this type of architecture proposed by \citet{caruana1993multitask}, the bottom layers of the neural network are shared among all the tasks, and the top layers are task-specific. On one hand, the main advantage of this class of methods is its scale invariance to a large number of tasks. On the other hand, with a shared representation the resulting features can become biased towards the tasks with strong signals.

The second group of MTL topologies have a dedicated set of parameters to each task. Such methods are called soft parameter sharing \cite{duong2015low}. They can be interpreted as single networks that have a feature sharing mechanism between encoders to induce inter branch information crossing. Methods like Cross-Stitch Network \cite{misra2016cross} and Multi-gate Mixture of Experts \cite{ma2018modeling} are examples of soft parameter sharing based on a explicit feature sharing mechanism, mixture of experts feature fusion and attention based approaches to cross-task among branches. Misra et al. \cite{misra2016cross} introduced the concept of soft-parameter sharing in deep multi-task approaches by learning a linear combination of the input activation maps. The linear combination (soft feature fusion) is learned at each layer from both tasks. The MMoE method proposed by \citet{ma2018modeling} is an attempt to provide a soft parameter sharing mechanic through a gating mechanism. The gate function selects a set of experts for each task while re-using it for multiple tasks, consequently providing feature sharing. The main advantage of soft parameter sharing approaches is the capability of learning task specific and shared representations explicitly. Nevertheless, these models suffer from scalability problems, as the size of the MTL network tends to grow proportionally with the number of tasks. A full survey on different MTL topologies can be found at \cite{ruder2017overview,vandenhende2020revisiting}.

The previously mentioned works focused on better network structures for MTL. Another significant problem of learning multiple tasks is related to the optimization procedure. MTL methods need to balance gradients of multiple tasks to prevent one or more tasks from dominating the network and producing task biased predictions. The optimization methods can be divided into loss balancing techniques \cite{liu2019loss, liu2019mtan}, gradient normalization \cite{chen2018gradnorm} and model-agnostic meta-learning \cite{finn2017model,lee2020multitask}. Liu et al. \cite{liu2019loss} proposed a loss balance approach based on loss ratios between the first batch and all subsequent ones in each epoch. The method called Loss-Balanced Task Weighting (LBTW) showed promising results reducing the negative transfer on a $128$ task scenario. Loss balancing approaches have as their main drawbacks its suboptimality when task gradients are conflicting or when a set of tasks have gradient magnitudes higher than others. In order to mitigate these limitations of loss based approaches GradNorm \cite{chen2018gradnorm} and Model-Agnostic Meta-Learning (MAML) for MTL \cite{finn2017model,lee2020multitask} were proposed.

Gradient normalization proposed by Chen et al. \cite{chen2018gradnorm} aims to control the training through a mechanism that encourages all tasks to have similar magnitude. Additionally to it the model also balances the pace tasks are learned. More recently, methods based on meta-learning \cite{finn2017model} emerged and outperformed previous loss-based approaches and gradient normalization techniques \cite{lee2020multitask}. Lee et al. \cite{lee2020multitask} proposed a multi-step approach which updates each task in an exclusive fashion. The method is capable of not only providing a balanced task optimization but also boosts current MTL architectures. MTL meta-learning methods, while being the current state-of-the-art class of approaches, can become impractical for settings with large cardinality based on the intermediate steps which are needed to task state computation.  

The proposed multi-gate mixture-of-experts with exclusivity (MMoEEx) approach has similarities with the MMoE approach \cite{ma2018modeling}  and MAML \cite{finn2017model}. Our technique is an MMoE approach, however we introduce an exclusivity mechanism that provides an explicit sparse activation of the network, enabling the method to learn task specific features and a shared representation simultaneously. We also tackle the scalabilty limitation of MMoE techniques with our exclusion gates. Our approach is inspired by the MAML technique in which we aim to introduce a two step approach to balance tasks at the gradient level for mixture of experts. 

\section{Methodology}\label{methodology}

Hard-parameter sharing networks \cite{caruana1993multitask} shown in Figure \ref{nnarchitectures}.b are one of the pillars of multi-task learning. These networks are composed of a shared bottom and task-specific branches. \citet{ma2018modeling} suggested that a unique shared bottom might not be enough to generalize for all tasks in an application, and proposed to use several shared bottoms, or what they call \textit{experts}. The experts are combined using gate functions, and their combination is forwarded to the towers. The final architecture is called Multi-gate Mixture-of-Experts (MMoE), and is shown in Figure \ref{nnarchitectures}.c. MMoE generalizes better than its traditional hard-parameter sharing counterpart, but there are two weaknesses: first, it lacks a task-balancing mechanism; second, the only source of diversity among the experts is due to the random initialization. Although the experts can indeed be diverse enough if they specialize in different tasks, there are no guarantees that this will happen in practice. In this work, we propose the Multi-gate Mixture-of-Experts with Exclusivity (MMoEEx) (Figure \ref{nnarchitectures}.d), a model that induces more diversity among the experts and has a task-balancing component. 

\subsection{Structure}

 The neural network architecture follows the structure proposed by \citet{ma2018modeling} and can be divided into three parts: gates, experts, and towers. Considering an application with $K$ tasks, input data 
$x \in \mathbb{R}^d$, the gate function $g^k()$ is defined as: 

\begin{equation}
\label{eqg}
    g^k(x) = softmax(W^k x), \forall k \in \{0,...,K\}
\end{equation}
where $W^k \in \mathbb{R}^{E \times d}$ are learnable weights and $E$ is the number of experts, defined by the user. The gates control the contribution of each expert to each task.

The experts $f_e(),\forall e\in \{0,...,E\}$, and our implementation is very flexible to accept several experts architectures, which is essential to work with applications with different data types. For example, if working with temporal data, the experts can be LSTMs, GRUs, RNNs; for non-temporal data, the experts can be dense layers. The number of experts $E$ is defined by the user. The experts and gates' outputs are combined as follows:
\begin{equation}\label{eqf}
    f^k(x) = \sum_{e=0}^Eg^k(x)f_e(x), \forall k \in \{0,...,K\}
\end{equation}

The $f^k()$ are input to the towers, the task-specific part of the architecture. Their design depends on the data type and tasks. The towers $h^k$ output the task predictions as follows: 
\begin{equation}
    y^k = h^k(f^k(x)), \forall k \in \{0,...,K \}
\end{equation}

\subsection{Diversity}

In ensemble learning, models with a significant diversity among their learners tend to generalize better. MMoE \cite{ma2018modeling} leverages several experts to make its final predictions; however, it relies only on random initialization to create diversity among the experts, and on the expectation that the gate function will learn how to combine these experts. Here we propose two mechanisms to induce diversity among the experts, defined as \textit{exclusion} and \textit{exclusivity}: 
\begin{itemize}
    \item Exclusivity: We set $\alpha E$ experts to be exclusively connected to one task. The value $\alpha\in[0,1]$ controls the proportion of experts that will be \textit{exclusive}. If $\alpha=1$, all experts are exclusive, and if $\alpha=0$, all experts are shared (same as MMoE). An exclusive expert is randomly assigned to one of the tasks $T_k$, but the task $T_k$ can still be associated with other exclusive experts and shared experts. 
    \item Exclusion: We randomly exclude edges/connections between $\alpha E$ experts and tasks. If $\alpha=1$, all experts will have one connection randomly removed, and if $\alpha=0$, there is no edge deletion (same as MMoE). 
\end{itemize}

For applications with only two tasks ($K=2$), exclusion and exclusivity mechanisms are identical. The exclusion mechanism is more scalable than the exclusivity mechanism because it does not require one expert per task, and therefore, works well in applications with a large number of tasks. For a small set of tasks, both approaches have similar results. MMoEEx, similarly to MMoE, relies on the expectation that gate functions will learn how to combine the experts. Our approach induces more diversity by forcing some of these gates to be `closed' to some experts, and the exclusivity and exclusion mechanisms are used to close part of the gates. The remaining non-closed gates learn to combine the output of each expert based on the input data, according to Equation \ref{eqg}.

As Equation \ref{eqf} shows, the gates are used as experts weights. Therefore, if an expert $e\in \{0,...,E\}$ is exclusive to a task $k \in \{0,...,K\}$, then only the value $g^k[e]\neq 0$, and all the other gates for that expert are `closed': $g^m[e] = 0, m\in \{0,...,K\}, m \neq k$.

\subsection{MAML - MTL optimization}

The goal of the two-step optimization is to balance the tasks on the gradient level. \citet{finn2017model} proposed the Model-agnostic Meta-learning (MAML), a two-step optimization approach originally intend to be used with transfer-learning and few-shot learning due to its fast convergence. MAML also has a promising future in MTL. \citet{lee2020multitask} first adapted MAML for Multi-task learning applications, showing that MAML can balance the tasks on the gradient level and yield better results than some existing task-balancing approaches. The core idea is that MAML's temporary update yields smoothed losses, which also smooth the gradients on direction and magnitude. 

In our work, we adopt MAML. However, differently from \citet{lee2020multitask}, we do not freeze task-specific layers during the intermediate/inner update. The pseudocode of our MAML-MTL approach is shown in Algorithm \ref{alg:maml1}.

\begin{algorithm}
\caption{MAML-MTL}
\label{alg:maml1}
Sample batch $X$\;
$loss$ = 0\; 
\For{$T$ in TASKS}{
Evaluate $\Delta_{\theta}\mathcal{L}_{T}(f_{\theta}(X))$ \;
Temporary Update $\theta'_T \leftarrow \theta - \Delta_{\theta}\mathcal{L}_{T}(f_{\theta}(X))$\; 
Re-evaluate and save $loss = loss+ \Delta_{\theta'_T}\mathcal{L}_{T}(f_{\theta'_T}(X))$\;
}
Update $\theta \leftarrow \theta - loss$\;
\end{algorithm}

\begin{figure}[!h]
    \centering
    \includegraphics[scale=0.7]{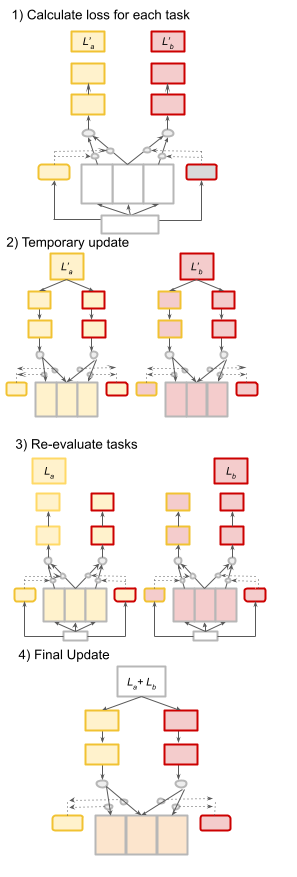}
    \caption{Illustration of MAML-MTL (Algorithm \ref{alg:maml1}) with two tasks. White blocks represent not task-specific layers; yellow and red represent task-specific blocks. Steps 2 and 3 show temporarily updated models (one for each task). Step 4 updates all the models with the sum of the two re-calculated losses.}
    \label{fig:maml}
\end{figure}

Figure \ref{fig:maml} shows an illustration of Algorithm \ref{alg:maml1} when we have two tasks ($A$ and $B$). The white boxes represent shared blocks of the neural network; yellow and red blocks are specific for task A and task B, respectively. The first step of MAML is to calculate the current loss for each task. A standard optimization method would normally sum these losses and back-propagate. However, these are the losses we want to smooth to improve task balancing. Therefore, we perform a temporary update in the entire neural network for each task, as shown in step 2. Then, in step 3, we re-evaluate tasks using the temporarily updated neural network. These new losses are smoothed (on magnitude and direction) and are used on the final update (step 4).

Section \ref{experiments} shows the results of the experiments using our two-step optimization strategy. One of the main drawbacks of this approach is scalibility. Temporary updates are expensive, making infeasible the use of MAML in applications with many tasks.

\section{Experiments}\label{experiments}

In this section, we develop experiments to answer two main questions to validate our proposed method: 
\begin{enumerate}
    \item Our proposed method MMoEEx has better results than existing MTL baselines, such as MMoE, Hard-parameter sharing (shared bottom), Multi-Channel Wise LSTMs (time-series datasets);
    \item Our proposed method MMoEEx, has better results than Single Task Learning (STL) methods. 
\end{enumerate}

Furthermore, we also explore some secondary results, such as the influence of the expert complexity and the number of experts on the results, and the comparison of expert's diversity in our proposed method and our main baseline.

\subsection{Datasets}

We evaluate the performance of our approach on three datasets. UCI-Census-income dataset \cite{ma2018modeling,dua2019uci}, Medical Information Mart for Intensive Care (MIMIC-III) database \cite{johnson2016mimic, harutyunyan2019multitask} and PubChem BioAssay (PCBA) dataset \cite{wang2017pubchem}. A common characteristic among all datasets is the presence of very unbalanced tasks (few positive examples). 

\subsubsection*{UCI - Census-income dataset\cite{ma2018modeling,dua2019uci}}
Extracted from the US 1994 census database, there are 299,285 answers and 40 features, extracted from the respondent's socioeconomic form. Three binary classification tasks are explored using this dataset: 
    \begin{enumerate}
        \item Respondent income exceeds \$50K; 
        \item Respondent's marital status is ``ever married''; 
        \item Respondent's education is at least college; 
    \end{enumerate}
    

\subsubsection*{Medical Information Mart for Intensive Care (MIMIC-III) database \cite{johnson2016mimic, harutyunyan2019multitask}}
\begin{figure}[!ht]
    \centering
    \includegraphics[scale = 0.6]{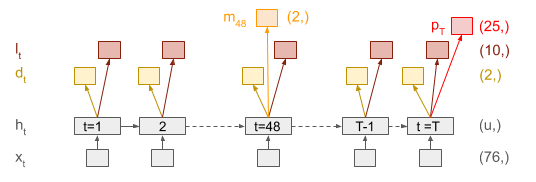}
    \caption{The neural network architecture adopted in our work is the same used by the Multi-Channel Wise LSTM\cite{harutyunyan2019multitask}. The input data $x_t$ has 76 features, and the size of the hidden layer $h_t$ depends on the model adopted. There are four tasks: the decompensation $d_t$ and LOS $l_t$ calculated at each time step, mortality $m_{48}$, and the phenotype $p_T$, both calculated only once per patient.}
    \label{mimic}
\end{figure}

This database was proposed by \citet{harutyunyan2019multitask} to be a benchmark dataset for MTL in time-series data. It contains metrics of patients from over 40,000 intensive care units (ICU) stays. This dataset has 4 tasks: two binary tasks, one temporal multi-label task, and one temporal temporal classification. Figure \ref{mimic} shows the neural network adopted in our work and where each task is calculated. 

We followed the data-preprocessing introduced by \citet{harutyunyan2019multitask}, available at \url{https://github.com/YerevaNN/mimic3-benchmarks}. To facilitate benchmarking, we kept the same tasks and tasks' structure. Tasks description: 
\begin{enumerate}
\item Phenotype prediction: measured on the end of stay, classify if the patient has 25 acute care conditions ($p_T$ in Figure \ref{mimic}). In practice, we have 25 parallel binary classification tasks; 
\item Length-of-stay (LOS) prediction: the goal is to predict the remaining time spend in ICU at each hour of stay ($l_t$ in Figure \ref{mimic}). Following \citet{harutyunyan2019multitask}, the remaining time was converted from a regression task to a multi-label task. We have 10 classes, one class for each one of the first 8 days, between 8-15 days, and +15 days;
\item Decompensation prediction: aim to predict if the patient state will rapidly deteriorate in the next 24 hours. We follow the \citet{harutyunyan2019multitask} scheme, where due to lack of gold standard, the task is redefined as mortality prediction in the next 24 hours at each hour of an ICU stay, in practice, a temporal binary classification ($d_t$ in Figure \ref{mimic}); 
\item In-hospital mortality prediction: binary classification in the end of the first 48 hours of a patient in an ICU stay ($m_{48}$ in Figure \ref{mimic}); 
\end{enumerate}

\subsubsection*{PubChem BioAssay (PCBA)\cite{wang2017pubchem, liu2019loss} Database} We worked with a subset of the PCBA, composed of 128 binary tasks/biological targets and 439,863 samples. Each sample represents a molecule, pre-processed using Circular Fingerprint \cite{liu2019loss} molecule feature extractor, that creates 1024 features.  These features are used to determine whether the chemical affects a biological target, here defined as our tasks.

\subsection{Design of experiments}

The split between train, validation, and test set was the same used by our baselines to offer a fair comparison. For the UCI-Census, the split was 66\%/17\%/17\% for training/validation/testing sets, for the MIMIC-III and PCBA 70\%/15\%/15\%. The data pre-processing, loss criterion, optimizers, parameters, and metrics description of our experiments is shown in Section \ref{Reproducibility}. The metric adopted to compare results is AUC (Area Under The Curve) ROC for the binary tasks and Kappa Score for the multiclass tasks. 

\subsection{Experiments Reproducibility}\label{Reproducibility}
\begin{table*}[]
    \centering
    \caption{Models' architecture, training information, and dataset pre-processing's references for experiment reproducibility.}
    \begin{tabular}{c|p{3cm}|c|c|p{3.5cm}|p{5.5cm}}
    \hline
    Dataset & Pre-processing &Epochs & Experts & Loss &Layers   \\ \hline
    UCI-Census   & \citet{ma2018modeling} & 200 & 12  & $BCEWithLogitsLoss$ & Experts: $Linear(4) + ReLU$, Towers: $Linear(4) + Linear(1)$\\\hline
    MIMIC-III & \citet{harutyunyan2019multitask,johnson2016mimic} & 50 & 12 and 32 & $BCEWithLogitsLoss$, $CrossEntropyLoss$ (multilabel task), $pos\_weight$: $pheno = 5$, $LOS=1$, $Decomp = 25$, $Ihm = 5$ &  Experts: RNN(128) or GRU(128); Towers: $Linear(16) + Linear(output)$, where the output depends on the task. Three towers had time-series data, and one had only the first 24 observations of the time-series.\\ \hline
    PCBA & \citet{liu2019loss} & 100& 2 or 4 & $BCEWithLogitsLoss$, $pos\_weight = 100$ &  $Linear(2000)+Dropout(0.25)+ReLU+Linear(2000)+Sigmoid+Linear(2000)$. The tower had one $Linear(1)$ layer per task. \\ \hline
    \end{tabular}
    \label{reproducibility}
\end{table*}

We used PyTorch in our implementation, and the code will be made available at \url{https://github.com/BorealisAI/MMoEEx-MTL}.
We used Adam optimizer with learning rate $0.001$, weight decay $0.001$, and learning rate decreased by a factor of $0.9$ every ten epochs. The metric adopted to compare the models was ROC AUC, with the exception of the task LOS on MIMIC-III dataset, which was Cohen's kappa Score, a statistic that measures the agreement between the observed values and the predicted. We trained the models using the training set, and we used the task's AUC sum in the validation set to define the best model, where the largest sum indicates the best epoch, and consequently, the best model. Table \ref{reproducibility} shows a summary of the models adopted for future reference.

\subsection{UCI - Census-income Study}

In this subsection, we report and discuss experimental results on the census-income data. We present experiments predicting income, marital status, and education level.

The census income, marital status and education dataset experiments are presented at Table \ref{census3tasks}. We can observe that our approach outperforms all the baselines with the exception of the Education task where the single-task method presents a marginal improvement over MMoEEx. We argue that the Census tasks already present slightly conflicting optimization goals and, in this situation, our approach is better suited to balance multiple competing tasks.

\begin{table}[ht]
\centering
\caption{Results on Census Income/Marital/Education dataset. $\Delta$ is the average relative improvement. MMoEEx Exclusivity $\alpha=0.5$.}
\begin{tabular}{c|c|c|c|c} \hline
\multirow{2}{*}{Method} & \multicolumn{3}{c|}{AUC} & \multirow{2}{*}{$\Delta$} \\ \cline{2-4}
 & Income & Marital Stat & Education& \\ \hline
 Single-Task & $88.95$ & $97.48$ & \textbf{87.23}& - \\ \hline
Shared-Bottom & $91.09$ & $97.98$ & $86.99$& $+0.85\%$ \\ \hline
MMoE  & $90.86$ & $96.70$ & $86.33$ & $-0.28\%$ \\ \hline
MMoEEx (Ours) & \textbf{92.51} & \textbf{98.47} & $87.19$& \textbf{+1.65\%} \\
\hline\end{tabular}
\label{census3tasks}
\end{table}

We can conclude that our MMoEEx approach can better learn multiple tasks when compared to standard shared bottom approaches and the MMoE baseline, due to the exclusivity and the multi-step optimization contributions of our work.

\subsection{MIMIC-III Study}

MIMIC-III dataset is the main benchmark for heterogeneous MTL with time series. The dataset consists of a 
mixture of multi-label and single-label temporal tasks and two non-temporal binary classification tasks. Our experiments will first investigate the best recurrent layers to be selected as experts to the MMoEEx model. The following sections show an ablation study on the impact of higher experts cardinality and our full scale baseline evaluation. 

\subsubsection{Recurrent Modules Ablation Study}

One of the main design choices for time series prediction is the type of recurrent unit to be deployed. The goal of this ablation study is to provide a thorough analysis on the impact of different recurrent layers to our approach. The layers taken into consideration range from the standard RNN's \cite{Rumelhart1986RNN}, LSTM's \cite{HochreiterLSTM} and GRU's \cite{cho2014gru} to modern recurrent layers like the Simple Recurrent Units (SRU) \cite{lei2018sru} and  Independent Recurrent Networks (IndRNN) \cite{li2018indrnn}.  

\begin{table}[!ht]
\centering
\caption{Results on MIMIC-III recurrent modules ablation study. All the MMoEEx configurations count with $12$ experts based on memory limitations of approaches like IndRNN and LSTM. MMoEEx Exclusivity $\alpha=0.5$.}
\begin{tabular}{c|c|c|c|c} \hline
Method & Pheno & LOS & Decomp & Ihm \\ \hline
MMoEEx - SRU \cite{lei2018sru} & $71.00$ & \textbf{57.88} & $96.67$ & $89.95$\\ \hline
MMoEEx - IndRNN \cite{li2018indrnn}& $67.49$ & $57.11$ & $95.89$ & \textbf{91.68}\\ \hline
MMoEEx - IndRNNV2 \cite{li2018indrnn} & $68.15$ & $54.48$ & $96.50$ & $90.58$\\ \hline
MMoEEx - LSTM \cite{HochreiterLSTM} & $73.48$ & $45.99$ & $96.54$ & $90.88$ \\ \hline
MMoEEx - RNN \cite{Rumelhart1986RNN} & $73.40$ & $55.56$ & $96.85$ & $91.08$\\ \hline
MMoEEx - GRU \cite{cho2014gru}& \textbf{74.08} & $54.48$ & \textbf{97.20} & $91.49$\\ \hline
\end{tabular}
\label{mimicrnns}
\end{table}

MIMIC-III recurrent modules ablation study is presented in Table \ref{mimicrnns}. SRU and  IndRNN outperform the other methods from length-of-stay (LOS) task. Besides that MMoEEx with IndRNN also is the top performer for the in-hospital mortality (Ihm) task. Besides the good performance of SRU and IndRNN for these tasks, they present an imbalanced performance over all considered tasks and also impose a memory and runtime burden, making the scalability of MMoEEx to higher number of experts infeasible. Taking the overall task performance into consideration, RNN and GRU outperform the compared recurrent approaches. RNN, in addition to being a top performer expert, also presented the lowest memory footprint and consequently is capable of providing MMoEEx with more experts if needed.        
From this part on we will have MMoEEx's with RNN's or GRU's as their recurrent layers.

\subsubsection{Impact of experts cardinality}

During the training of our approach for the MIMIC-III experiments we noticed that a larger number of experts, when connected with our exclusivity mechanism, gave better overall results. In order to further explore this parameter we conducted a series of experiments where we trained our MMoEEx with RNN's with a number of experts ranging from $12$ to $64$ experts. We choose RNN's as the recurrent layer in this experiment based on its low memory requirement. 

\begin{figure}[!ht]
    \centering
    \includegraphics[scale=0.38]{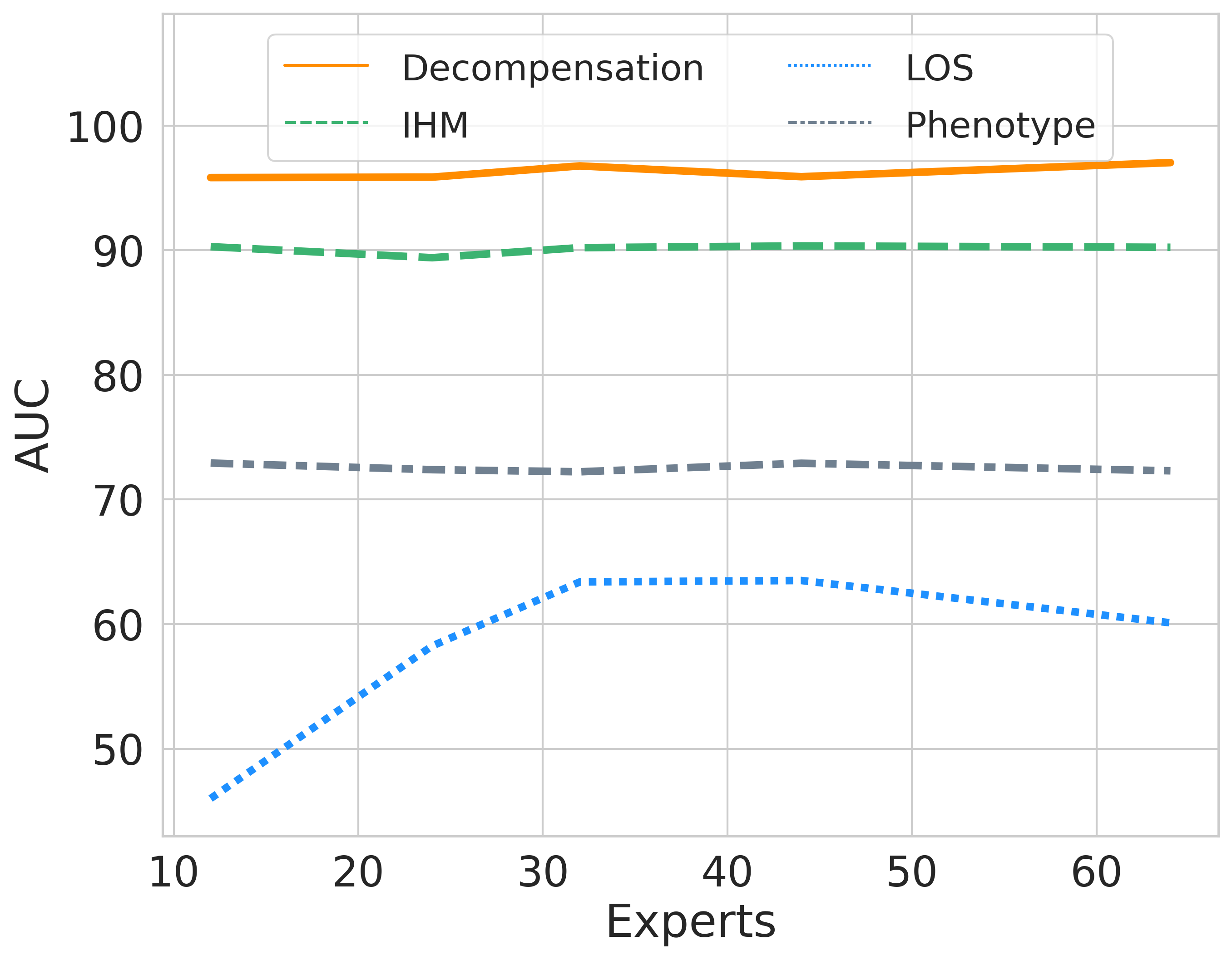}
    \caption{Impacts of cardinality of experts for MMoEEx model on MIMIC-III.}
    \label{abl:experts}
\end{figure}

Figure \ref{abl:experts} depicts our results for the four tasks on the MIMIC-III dataset. LOS tasks is the one that take most advantage of a larger number of experts with an improvement superior to $17$ percentage points or  a $38$ percent relative improvement. The remaining tasks are stable for a higher cardinality of experts. We believe a higher number of experts allow MMoEEx to have a better representation to challenging tasks when the shared representation is not been updated with the same magnitudes due to the other tasks have reached stability. The number of $32$ experts gave MMoEEx the best overall and LOS performance. The final results on MIMIC-III are all using $32$ experts.

\begin{table}[!ht]
\centering
\caption{Final results MIMIC-III. MMoEEx outperforms all the compared baselines with the exception to Pheno where the work from \citet{johnson2016mimic} is the best approach. Our approach can provide a relative improvement superior to $40$ percentage points when compared to the Multitask channel wise LSTM for the LOS task. MMoEEx Exclusivity $\alpha=0.5$.}
\begin{tabular}{c|c|c|c|c|c} \hline
Method & Pheno & LOS & Decomp & Ihm & $\Delta$ \\ \hline
MCW-LSTM \cite{johnson2016mimic} & \textbf{77.4} & $45.0$ & $90.5$ & $87.0$ & $+0.28\%$\\ \hline
Single Task \cite{johnson2016mimic} & $77.0$ & $45.0$ & $91.0$ & $86.0$ & -\\ \hline
Shared Bottom & $73.36$ & $30.60$ & $94.12$ & $82.71$ & $-9.28\%$ \\ \hline
MMoE & 75.09 & $54.48$ & $96.20$ & $90.44$ & $+7.36\%$\\ \hline
MMoEEx - RNN  & $72.44$ & \textbf{63.45} & $96.82$ & $90.73$ & \textbf{+11.74\%}\\ \hline
MMoEEx - GRU & $74.57$ & $60.63$ & \textbf{97.03} & \textbf{91.03} & \textbf{+11.00\%} \\ \hline
\end{tabular}
\label{mimic_final}
\end{table}

\subsubsection{MiMIC-III Results}

The full set of results for MIMIC-III dataset is presented in Table \ref{mimic_final}. We compared our approach with the multitask channel wise LSTM (MCW-LSTM) \cite{johnson2016mimic}, single task trained network, shared bottom, MMoE \cite{ma2018modeling} and two variations of MMoEEx with RNN's and GRU's.

 MMoEEx outperforms all the compared approaches except on the Phenotype (Pheno) task. For both time series tasks (LOS and Decomp) our approach outperforms all baselines. It is worth noting that for the LOS task, which is the hardest task on MIMIC-III, we present a relative improvement superior to $40$ percentage points when compared to multitask channel wise LSTM \cite{johnson2016mimic} and  over $16$ percentage points to MMoE for our MMoEEx with RNN's. MMoEEx with GRU's presents a better individual task performance than its RNN counterpart but with lower LOS task performance.

\subsection{PubChem BioAssay Dataset Study}

\begin{table}[!ht]
    \centering
    \small
    \caption{PCBA's final results. Our approach has competitive results when compared with the baselines. NT is Negative Transfer, $\Delta$ is Average Relative Improvement. 
    }\label{pcba_tab}

    \begin{tabular}{l|c|c|c|c}
     \hline
     \multirow{ 2}{*}{Method} & Average & CI $(95\%)$ & NT & $\Delta$ \\
     &  AUC &  & & \\ 
    \hline
    STL\cite{liu2019loss}           & 79.9 &  [78.0, 81.7] & - & -\\
    MTL\cite{liu2019loss}           & 85.7 &  [84.2, 87.2]  & 13& +8.5\%\\
    Fine Tuning \cite{liu2019loss}  & 80.6 &  [78.7, 82.4] & 50&+0.8\% \\
    GradNorm\cite{liu2019loss}      & 84.0  &  [82.5, 85.3] & 44& +5.1\%\\
    RMTL\cite{liu2019loss}                   & 85.2  &  [83.7, 86.7] & 11 &+6.6\%\\
    LBTW($\alpha=0.1$)\cite{liu2019loss}     & 85.9  &  [84.4, 87.3] & 13 &+7.5\%\\
    LBTW($\alpha=0.5$)\cite{liu2019loss}     & 86.3 &  [84.8, 87.6] & 11 &+8.0\%\\
    Shared Bottom   & 86.8 &  [84.6, 87.5] & 10 &+8.6\%\\
    MMoE            & 85.8  &  [84.1, 87.1] &15 &+7.3\%\\
    MMoEEx (ours)   & 85.9 &   [84.1, 87.1] & 13&+7.5\% \\ \hline
    \end{tabular}
\end{table}

The PCBA dataset has 128 tasks and is the main benchmark for scalability and negative transfer.  All the 128 tasks are binary classification tasks, and they are very similar to each other. Our experiments first compare our approach with existing baselines on the tasks' average AUC and number of tasks with negative transfer. Then, we have a second ablation study comparing our MMoEEx approach with the MMoE on the number of experts and overfitting evaluation. 

\subsubsection{Comparison with existing baselines}

 We used the work of \citet{liu2019loss}, as our main baseline, due to the variety of tested technique, such as GradNorm \cite{chen2018gradnorm}, and LBTW \cite{liu2019loss}. Additionally we included a shared bottom and a MMoE techniques to our baselines. 

\begin{figure}[!ht]
    \centering
    \includegraphics[scale=0.45]{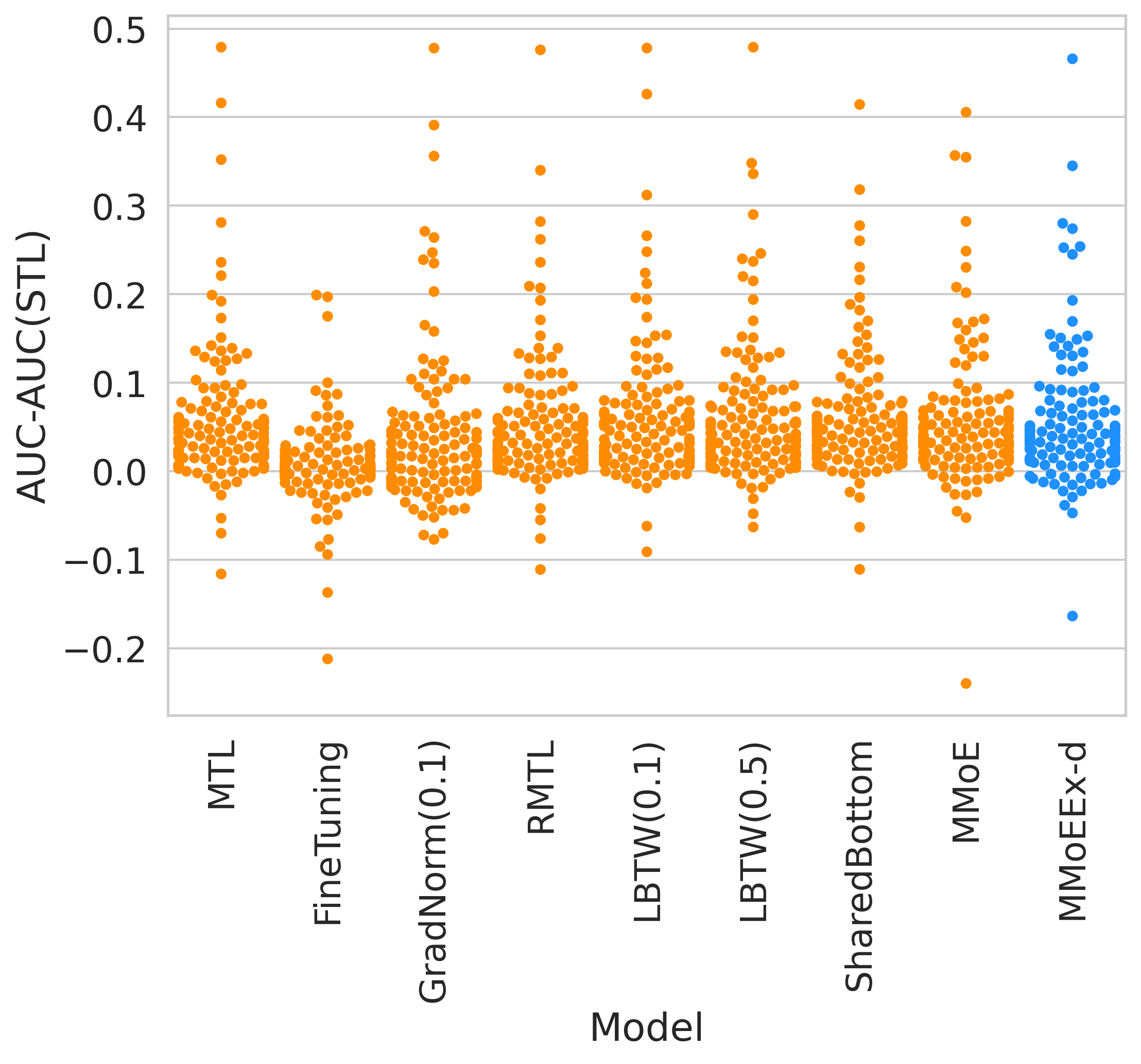}
    \caption{The plot shows change on the AUC for each task $k\in\{1,...,128\}$ relative to the Single-task Learning(STL) AUC. Values below 0 indicate negative transfer. MMoEEx Exclusivity $\alpha=0.5$.}
    \label{pcba_baselines}
\end{figure}

The architecture adopted for baselines and experts is very similar (more details see Section \ref{Reproducibility})
For this application, we did not use MAML-MTL optimization due to scalability issues. 
Therefore, the difference between the MMoE and MMoEEx in this application is the diversity of experts: all MMoE's experts are shared among all tasks, \textit{versus} only a portion of MMoEEx are shared.  Table \ref{pcba_tab} shows the final results. We adopted four metrics to compare our results with the baselines: the average ROC AUC of all tasks, Standard Deviation of the ROC AUC, $\Delta$, and the number of negative transfer (NT). The NT is calculated using Single Task Learning Models, and counts how many tasks have a worse result on the multi-task learning approach. Figure \ref{pcba_baselines} shows the improvement of each model in comparison with the STL model, where tasks below 0 indicates NT.

Considering all the baselines, the shared bottom fitted in our study has the best overall result (largest average AUC, smaller NT). Using the tasks AUC, we constructed 95\% confidence intervals, shown in Table \ref{pcba_tab}, from where we conclude there is no significant difference between RMTL, MTL, LBTW, Shared Bottom, MMoE, and MMoEEx. Therefore, our proposed method MMoEEx has a competitive result when compared with other baselines. We highlight that LBTW and GradNorm are both focused on task balancing. However, the PCBA dataset has very similar tasks, which almost makes unnecessary the task balancing component. The shared bottom model, for example, does not have any task balancing approach and has the best performance overall. 

\subsubsection{Impact of number of experts}

\begin{figure}[!ht]
    \centering
    \includegraphics[scale=0.45]{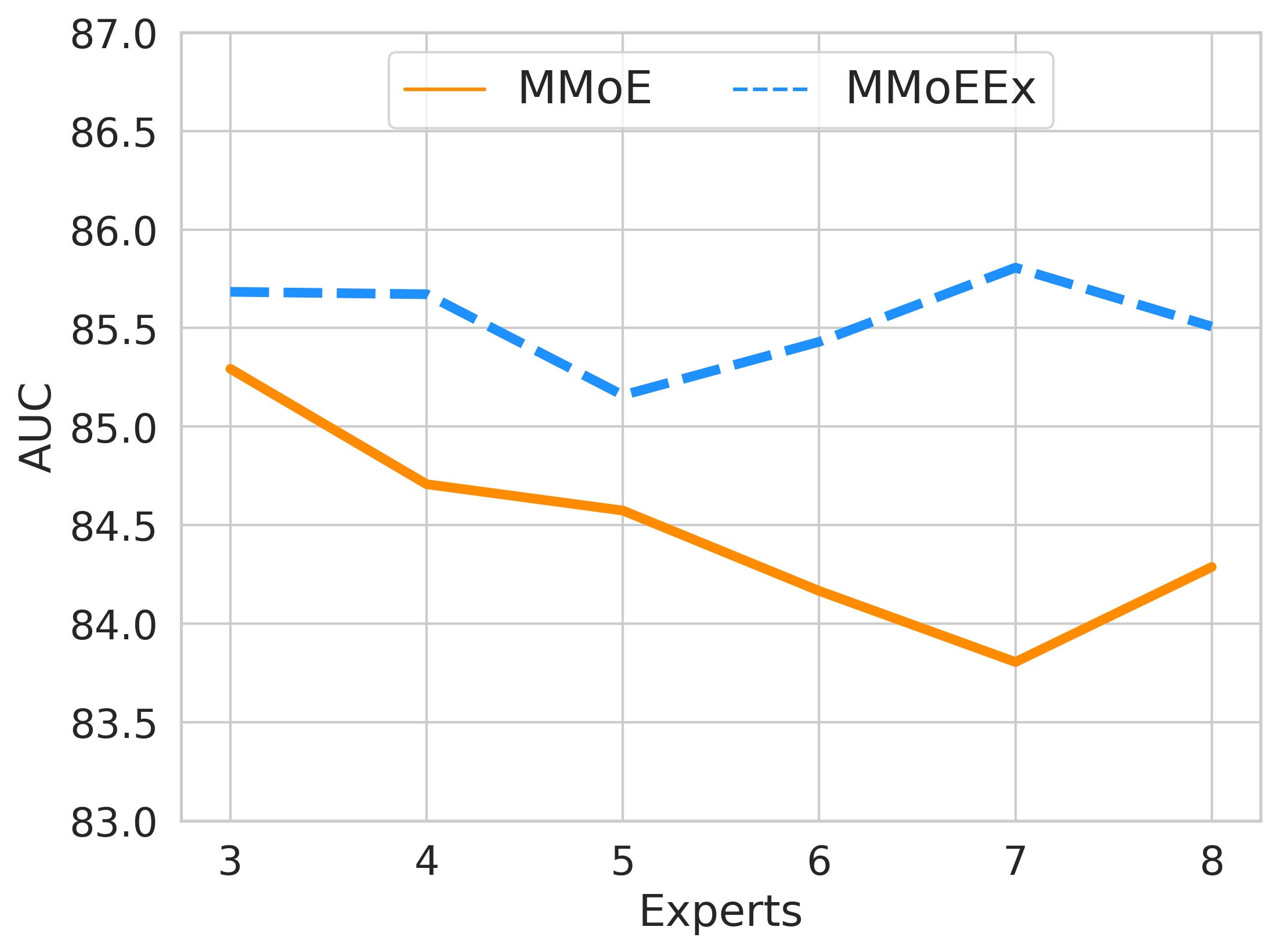}
    \caption{Comparison of the ROC AUC \textit{versus} number of experts. We changed the MMoEEx exclusivity parameter $\alpha$ by increments of 0.09, starting with $\alpha=0.42$, so there are always only two shared experts.}
    \label{pcba_experts}
\end{figure}

\begin{figure*}[!ht]
    \centering
    \includegraphics[scale=0.65]{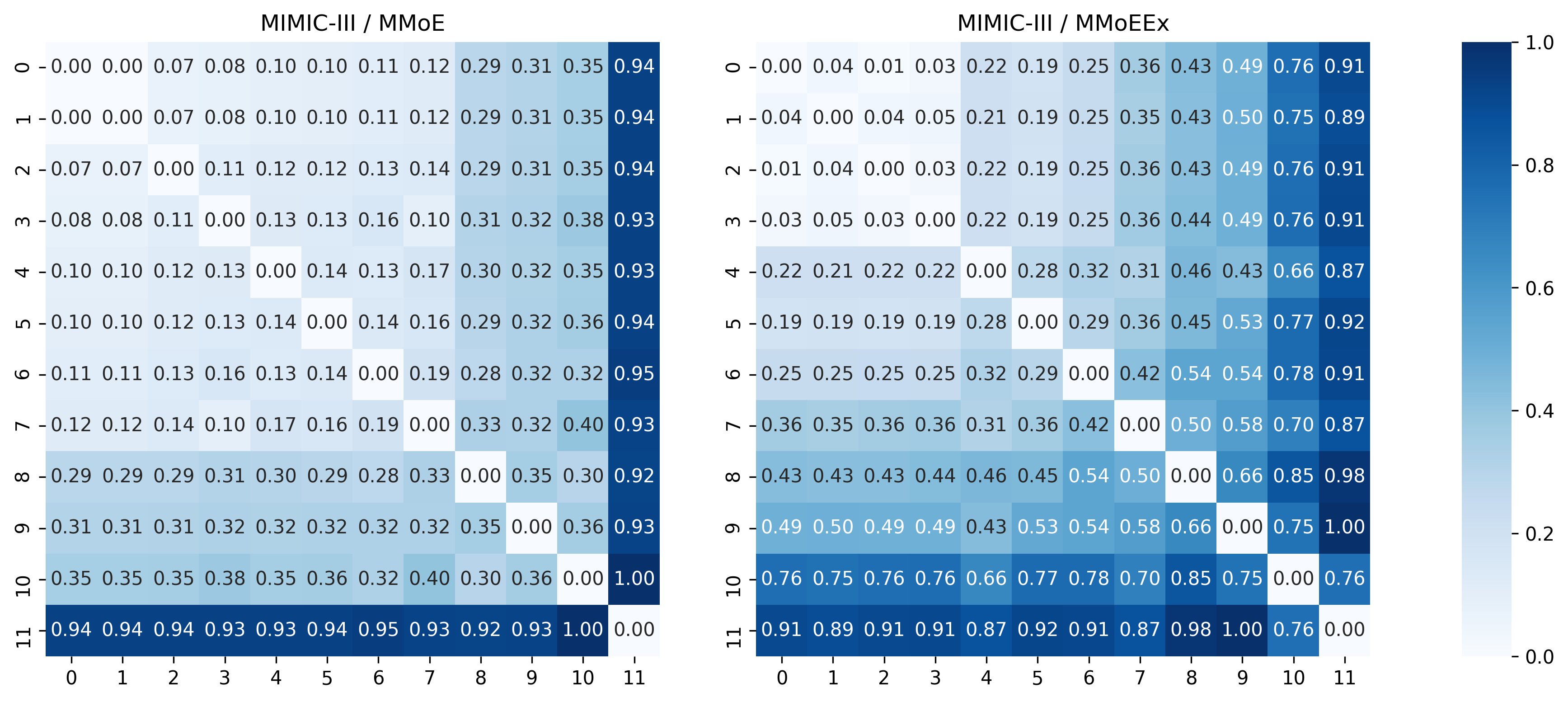}
    \caption{MMoE ($\bar{d}=0.311$) and MMoEEx($\bar{d}=0.445$) heatmap in the MIMIC-III dataset. The MMoE has 12 shared experts \textit{versus} 6 shared and 6 exclusive experts in the MMoEEx model. Darker colors indicate more dissimilarities between two experts and, therefore, more diversity. The plot is generated with 12 instead of 32 experts to better visualize the distances; the results also hold in the setting with 32 experts.}
    \label{diversity_mimic}
\end{figure*}

We also make a direct comparison between our method MMoEEx and our main baseline MMoE. In this dataset, fixing the number of experts, our proposed method MMoEEx has a better average ROC AUC on the testing set than the MMoE, as Figure \ref{pcba_experts} shows. In this study, we fix the number of shared experts in the MMoEEx as 2. With three experts, we adopted $\alpha=0.42$, and to each new expert added, we increment the value of $\alpha$ by 0.09. Therefore, with eight experts, we have two shared experts and $\alpha = 0.87$. The plot shows that the inclusion of more diversity on the experts through expert exclusivity helped the model to generalize better on the testing set and decreased overfitting.

\subsection{Diversity Score Study}

We propose a diversity measurement to support our claim that our method MMoEEX induced more diversity among the experts than our baseline MMoE. The diversity among the experts can be measured through the distance between the experts' outputs $f_e, \forall e\in\{0,..., E\}$. Considering a pair of experts $i$ and $j$, the distance between them is defined as:
\begin{equation}
    d_{i,j} = \sqrt{\sum_{n=0}^N(f_i(x_n)-f_j(x_n))^2}
\end{equation}
where $N$ is the number of samples in the dataset, $d_{i,j} = d_{j,i}$, and a matrix $D \in \mathbb{R}^{E\times E}$ is used to keep all the distances. To scale the distances into $d_{i,j}\in [0,1]$, we divide the raw entries in the distance matrix $D$ by the maximum distance observed, $max(D)$. A pair of experts $i,j$ with $d_{i,j} = 0$ are considered identical, and experts distances $d_{i,j}$ close to 0 are considered very similar; analogously, experts with $d_{i,j}$ close to 1 are considered very dissimilar. To compare the overall distance between the experts of a model, we define the \textit{diversity score} $\bar{d}$ as the mean entry in $D$. 

\begin{figure}[!ht]
    \centering
    \includegraphics[scale=0.55]{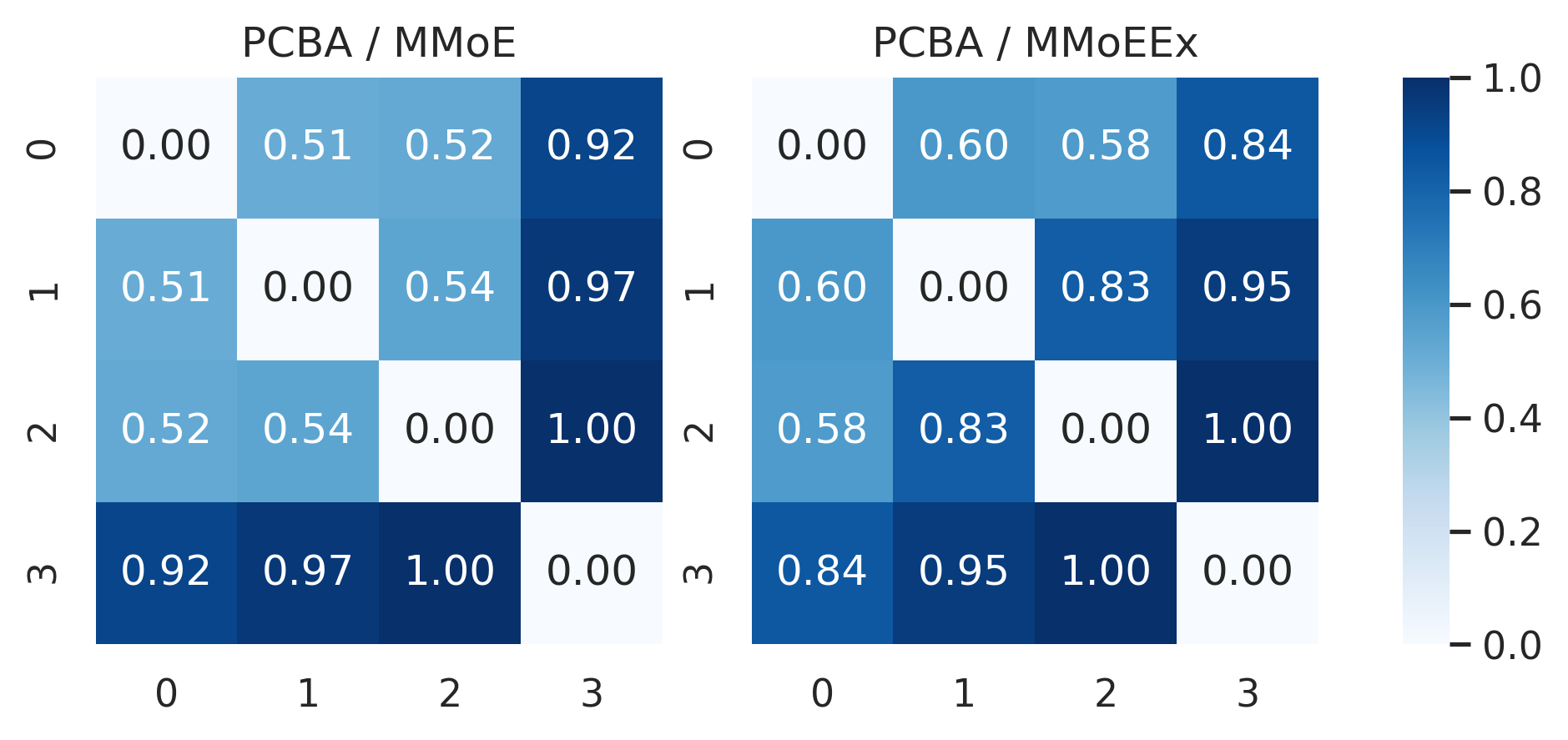}
    \caption{MMoE ($\bar{d} = 0.557$) and MMoEEx($\bar{d} = 0.600$) heatmap for the PCBA dataset. The MMoEEx model has 2 shared experts and 2 experts with exclusion. }
    \label{diversity_pcba}
\end{figure}

In this section, we analyze the diversity score of the MMoE and MMoEEx in our benchmark datasets.  The MMoE and MMoEEx models compared using the same dataset have the same neural network structure, but the MMoEEx uses the MAML - MTL optimization and has the diversity enforced. The MMoEEx models in Figures \ref{diversity_mimic} and \ref{diversity_pcba} were created with $\alpha = 0.5$ and exclusivity. In other words, half of the experts in the MMoEEx models were randomly assigned to be exclusive to one of the tasks, while the MMoE results have $\alpha = 0$  (all experts shared among all tasks). Figure \ref{diversity_mimic} shows a heatmap of the distances $D^{MMoE}$ and $D^{MMoEEx}$ calculated on the MIMIC-III testing set with 12 experts. The MMoE's heatmap has, overall, lighter colors and a smaller diversity score than the MMoEEx. Figure \ref{diversity_pcba} shows the MMoE and MMoEx heatmap for the PCBA dataset, with 128 tasks and 4 experts. Our proposed method MMoEEx also has a larger diversity score $\bar{d}$ and darker colors. 

In summary, our approach works extremely well on the heterogeneous dataset, MIMIC-III, increasing the diversity score by $43.0$\%. The PCBA is a homogeneous dataset, but the diversity component still positively impacts and increases the diversity score by $7.7$\%. Finally, as the most homogeneous and simplest dataset adopted in our study, the Census dataset is the only one that does not take full advantage of the experts' diversity. MMoE's diversity score was $0.410$ \textit{versus} $0.433$ for the MMoEEx's model, which is only $5.6$\% improvement. 

These results show that our approach indeed increased the experts' diversity while keeping the same or better tasks' AUC (See Tables \ref{census3tasks}, \ref{pcba_tab} and \ref{mimic_final}), which was one of our main goals.






\subsection{Biological Implications}

An MTL learning approach can improve generalization when the tasks are related from the ML standpoint. From a biological point of view, it means related tasks benefit from being modeled together, even when they are very distinct tasks. The explanation is that modeling multiple tasks in a single model can help the neural network to remove spurious correlations, highlighting important patterns within the data. To illustrate this intuition, consider the PCBA database and its results shown in Table \ref{pcba_tab}. Our goal is to predict if a given molecule will react to a chemical compound. The molecules are pre-processed using the circular fingerprint methodology, which creates a representation of the molecule structure using the atom neighborhoods. The final representation is a flat array with 1024 features for each molecule. Using an STL approach, we would have one model for each chemical compound; for each model, we would split the database into training/testing sets and train the model. Ideally, the model would learn to identify a representation of the input features, which is meaningful for predicting the observed reaction. That would be repeated for each one of the 128 chemical compounds, meaning that the knowledge acquired by predicting one compound under an STL approach is never re-used or improved over time. Every time, the STL model would learn everything from scratch. If a specific compound is unbalanced on the training set or harder to learn because it has a more complex behavior, the neural network might fail to capture and learn a meaningful representation adequately or give too much weight to spurions correlations. An MTL learning approach, on the other hand, would attempt to learn a meaningful representation of all chemical compounds, and the task-specific layers would then learn a specific representation for a given compound. That means the MTL approach is more likely to learn a better representation of the data because there is more information (from multiple tasks/losses) to help rule out spurious correlations and highlight important patterns. Yet, the task-specific layers allow the neural network to differentiate the tasks, giving room to compound specific representation. That intuition is confirmed by our experiments, as Table \ref{pcba_tab} shows: all MTL learning approaches had a better performance than the STL models.

\section{Conclusion}

We proposed a novel multi-task learning approach called Multi-gate Mixture-of-Experts with Exclusivity (MMoEEx), which extends MMoE methods by introducing an exclusivity and exclusion mechanism that induces more diversity among experts, allowing the network to learn representations that are more effective for heterogeneous MTL. We also introduce a two-step optimization approach called MAML-MTL, which balances tasks at the gradient level and enhances MMoEEx's capability to optimize imbalanced tasks. We show that our method has better results than baselines in MTL settings with heterogenous tasks, which are frequently found in biological applications. Experiments on biological and clinical benchmark datasets demonstrate the success of our proposed method in homogeneous and heterogeneous settings, where we outperform several state-of-the-art baselines.

Besides the contributions above, we believe the ablation analysis on MMoEEx can open the opportunity to further explore heterogeneous multi-task learning. The substantial improvement on the hardest task of the MIMIC-III dataset is an indication of this possible investigation direction. The MAML-MTL optimization is also in its infancy, and more research on meta-learning task balancing can greatly benefit MTL research. We hope this work inspires other researchers to further investigate multi-task learning at the network architecture and optimization levels.

\ifCLASSOPTIONcaptionsoff
  \newpage
\fi



%

\bibliographystyle{IEEEtranN}

\bibliography{references}

%

\begin{IEEEbiography}[{\includegraphics[width=1in,height=1.25in,clip,keepaspectratio]{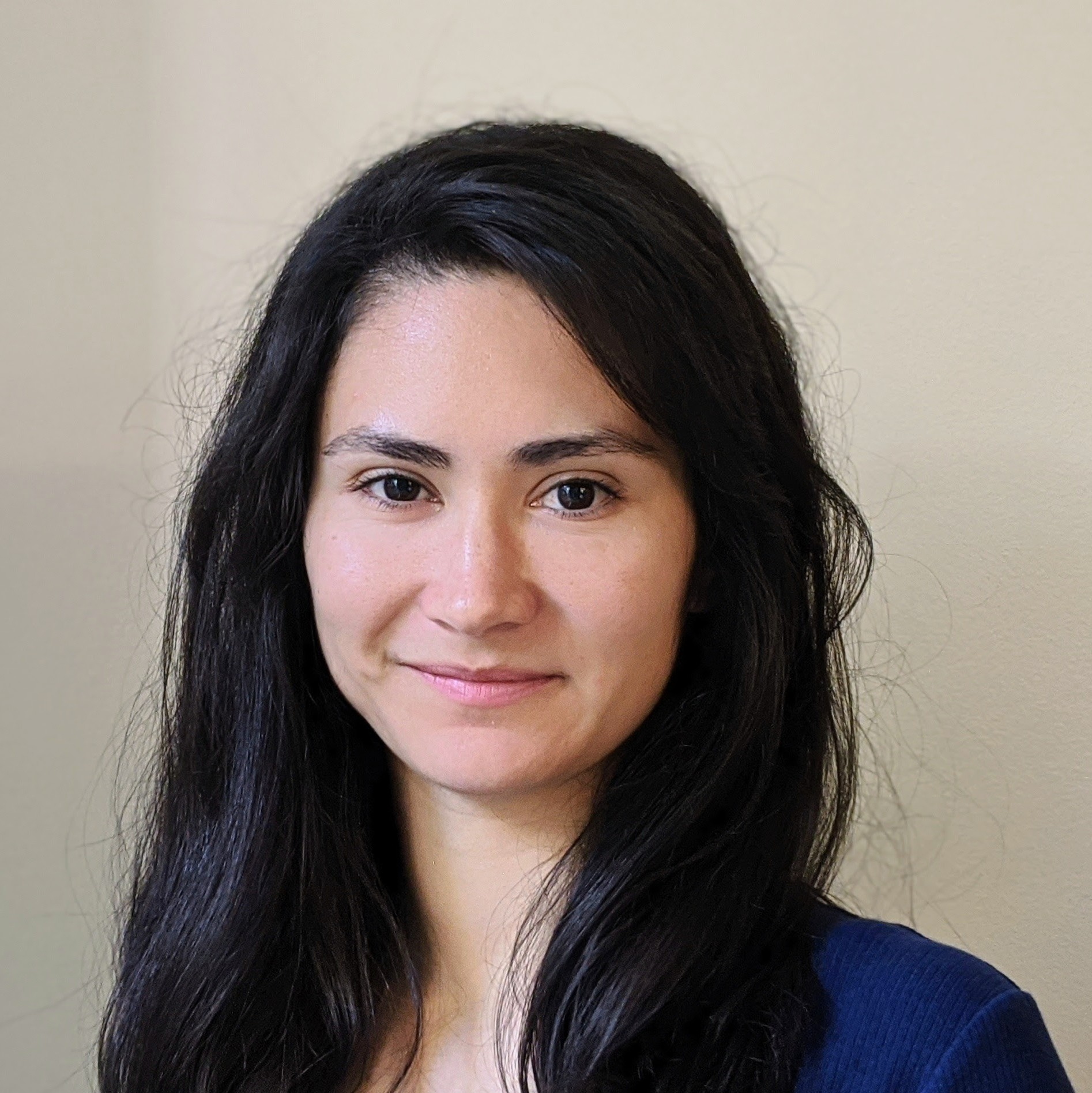}}]{Raquel Aoki}
 received a B.S. in Statistics and M.S. in Computer Science from the Federal University of Minas Gerais in 2014 and 2017, respectively. She is currently pursuing a Ph.D. degree in Computer Science at Simon Fraser University, Canada.  Her researcher interests include causality for computational biology and multi-task learning. This work was developed during her internship at Borealis AI.
\end{IEEEbiography}

\begin{IEEEbiography}[{\includegraphics[width=1in,height=1.25in,clip,keepaspectratio]{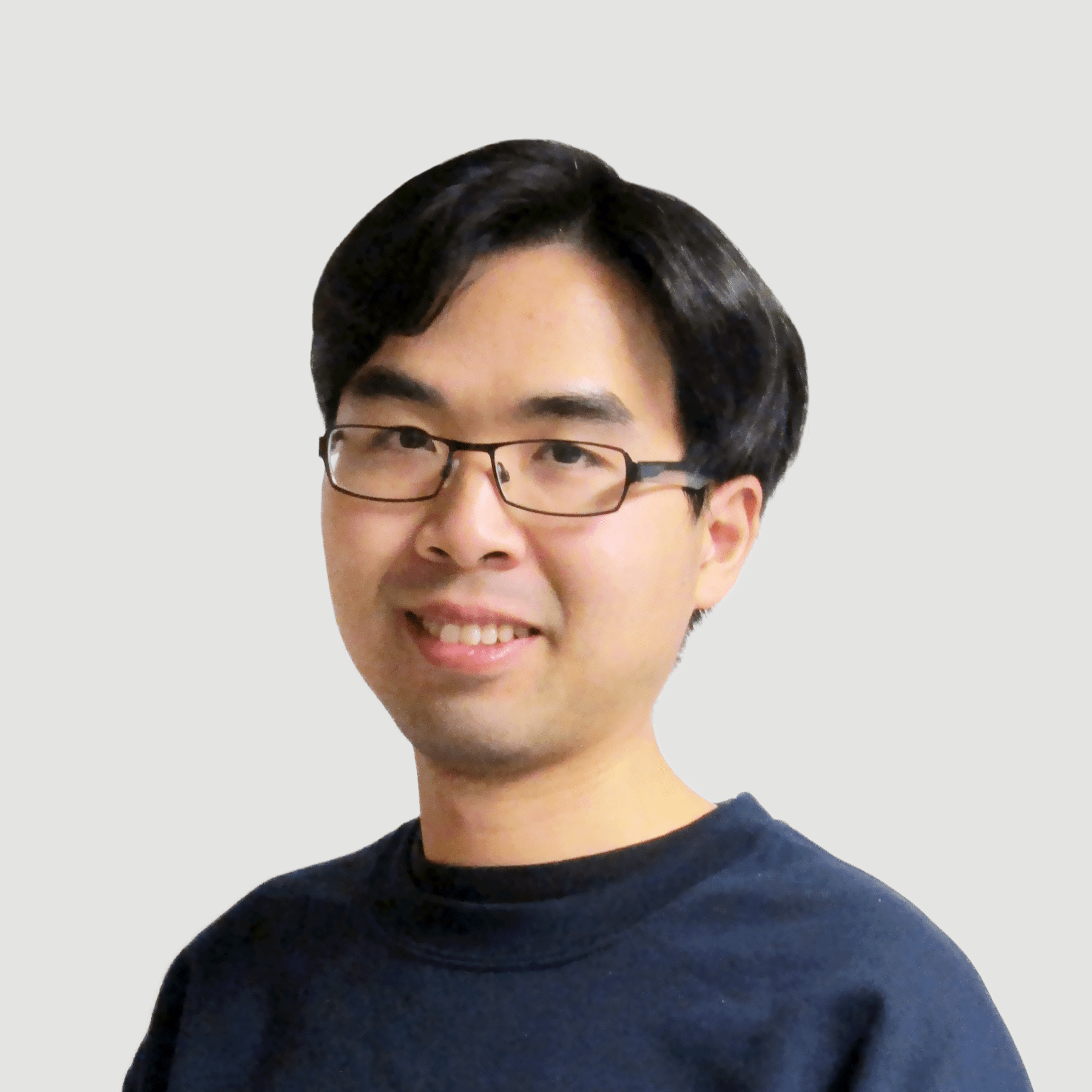}}]{Frederick Tung}
 received the Ph.D. degree in computer science from the University of British Columbia in 2017 and is currently a research team lead with Borealis AI. His research interests are in computer vision and resource-efficient deep learning.
\end{IEEEbiography}

\begin{IEEEbiography}[{\includegraphics[width=1in,height=1.25in,clip,keepaspectratio]{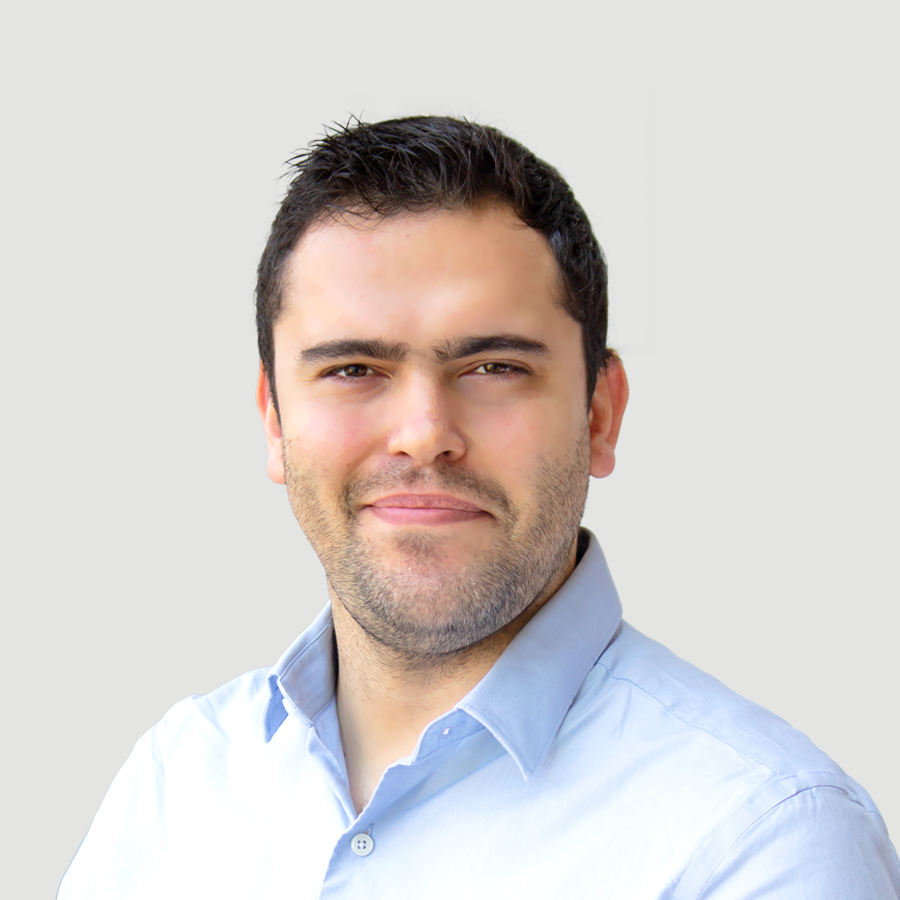}}]{Gabriel L. Oliveira}
 received a B.S. in Computer Engineering from the Federal University of Rio Grande in 2009, a M.S. in Computer Science from the Federal University of Minas Gerais and University of Minnesota in 2012 and 2014, respectively. He received a Ph.D. in Computer Science from the University of Freiburg in 2019. In the same year he joined Borealis AI, as a machine learning researcher. His research interests include multi-task learning, long-tail distribution learning and dynamic neural networks.
\end{IEEEbiography}




\end{document}